\documentclass{article}

\PassOptionsToPackage{numbers, compress}{natbib}
\PassOptionsToPackage{dvipsnames,svgnames}{xcolor}
\usepackage[preprint]{arxiv}

\usepackage{amsmath,amsfonts,bm}

\def\eqref#1{equation~\ref{#1}}

\def\1{\bm{1}}

\DeclareMathAlphabet{\mathsfit}{\encodingdefault}{\sfdefault}{m}{sl}
\SetMathAlphabet{\mathsfit}{bold}{\encodingdefault}{\sfdefault}{bx}{n}

\usepackage[utf8]{inputenc}
\usepackage[T1]{fontenc}
\usepackage{hyperref}
\usepackage{url}
\usepackage{booktabs}
\usepackage{nicefrac}
\usepackage{microtype}
\usepackage[dvipsnames,svgnames]{xcolor}
\usepackage{colortbl}
\usepackage{listings}
\usepackage{graphicx}
\usepackage{makecell}
\usepackage{wrapfig}
\usepackage{caption}
 \usepackage{pifont}
 \usepackage{amsmath}
\usepackage{algorithm}
\usepackage{algpseudocode}
\usepackage{fontawesome}
\usepackage{marvosym}

\makeatletter
\newcommand*{\myfnsymbol}[1]{%
  \ensuremath{%
    \ifcase#1\or
      \text{\ding{41}}\or
      1\or
      2\or
      3\or
      4\or
      5\or
      6\or
      7\else
      8\fi
  }%
}
\def\@fnsymbol#1{\myfnsymbol{#1}}
\def\thempfootnote{\myfnsymbol{\c@mpfootnote}}
\makeatother

\hypersetup{
    colorlinks=true,
    linkcolor=RoyalBlue,
    citecolor=ForestGreen,
    urlcolor=NavyBlue,
    anchorcolor=black
}

\usepackage{multirow}
\usepackage{ulem}

\useunder{\uline}{\ul}{}
\newcommand{\firsttablegroup}[2]{\multicolumn{#1}{@{}l}{\textcolor{black!55}{\itshape\footnotesize #2}}\\[-0.15em]}
\newcommand{\tablegroup}[2]{\midrule\multicolumn{#1}{@{}l}{\textcolor{black!55}{\itshape\footnotesize #2}}\\[-0.15em]}
\newcommand{\circnum}[1]{\raisebox{0.15ex}{\scriptsize\textcircled{\raisebox{-0.05ex}{\tiny #1}}}}

\definecolor{myred}{RGB}{228, 55, 62}
\definecolor{mygreen}{RGB}{14, 134, 114}

\lstdefinestyle{pythoncode}{
    language=Python,
    basicstyle=\ttfamily\footnotesize,
    keywordstyle=\color{RoyalBlue},
    commentstyle=\color{ForestGreen},
    stringstyle=\color{BrickRed},
    showstringspaces=false,
    columns=fullflexible,
    keepspaces=true,
    breaklines=true,
    frame=single,
    rulecolor=\color{gray!50},
    xleftmargin=0.5em,
    xrightmargin=0.5em,
    framexleftmargin=0.5em,
    framexrightmargin=0.5em,
    captionpos=b
}

\algblock[Block]{With}{EndWith}

\title{High-Fidelity Two-Step Image Generation via Teacher-Aligned End-to-End Distillation}

\author{
\textbf{Dongyang Liu}$^{1,2}$\textsuperscript{\ensuremath{\dagger}} \quad
\textbf{Ruoyi Du}$^1$ \quad
\textbf{David Liu}$^2$\textsuperscript{\ensuremath{\dagger}} \quad
\textbf{Dengyang Jiang}$^1$ \quad
\textbf{Liangchen Li}$^1$\\
\textbf{Qilong Wu}$^1$ \quad
\textbf{Zhen Li}$^1$ \quad
\textbf{Steven C.H. Hoi}$^1$ \quad
\textbf{Hongsheng Li}$^2$\textsuperscript{\ding{41}} \quad
\textbf{Peng Gao}$^1$\thanks{Corresponding authors.}\\[1mm]
$^1$Z-Image Team, Alibaba Group \quad
$^2$The Chinese University of Hong Kong
}

\begin{document}

\maketitle
\begingroup
\renewcommand{\thefootnote}{\ensuremath{\dagger}}\footnotetext{Work done during an internship at Z-Image Team, Alibaba Group.}
\endgroup

\begin{abstract}
Few-step diffusion distillation has become increasingly mature for 4--8-step generation, yet pushing further to 2 steps remains challenging. In this work, we introduce \textbf{Z-Image Turbo++}, a high-quality 2-step image generation model distilled from the 8-step Z-Image Turbo teacher. Our method addresses the central bottlenecks of the increased task difficulty and limited model capacity in 2-step generation through three simple but effective design choices tailored to this regime. First, we propose \textit{Distribution-Aligned Adversarial Learning}, which uses teacher-generated images rather than external real images as real samples for GAN training, providing a more attainable and informative adversarial target. Second, we adopt \textit{Step-Decoupled Parameterization}, assigning independent model parameters to the two denoising steps to better match their distinct capacity demands. Third, we perform \textit{End-to-End Training with Iterative Regularization}, allowing the first step to receive gradients from final image quality while preserving a meaningful intermediate generation through an explicit step-1 loss. Together, these designs substantially narrow the quality gap between 2-step and 8-step generation in both qualitative and quantitative evaluations, highlighting the potential of carefully tailored distillation strategies for improving the quality--efficiency trade-off in few-step generation.
\end{abstract}

\begin{figure}
    \centering
    \includegraphics[width=1\linewidth]{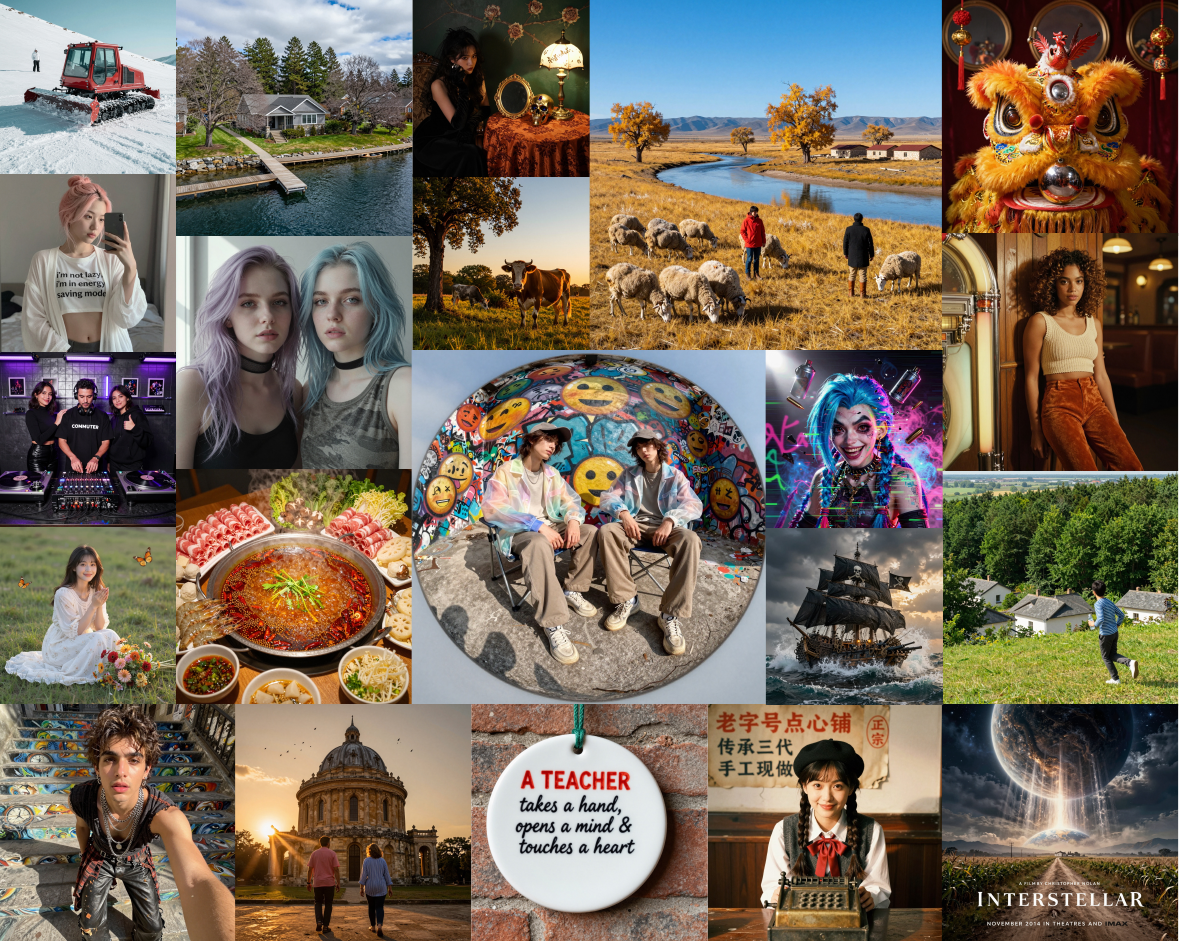}
    \caption{Images generated by Z-Image Turbo++ with only 2 steps. Best viewed with zoom.}
    \label{fig:overview}
\end{figure}
\section{Introduction}
\label{sec:intro}

Diffusion models~\citep{sohl2015deep, ho2020denoising, song2020score} have achieved remarkable success in image generation, producing outputs of exceptional quality and diversity. However, this performance comes at a significant computational cost: the iterative sampling process typically requires 40--100 neural network evaluations, creating a substantial barrier for deployment.

Few-step distillation has emerged as the dominant paradigm for addressing this bottleneck. Methods such as Distribution Matching Distillation (DMD)~\citep{dmd,diffinstruct,decoupleddmd}, Consistency Models~\citep{song2023consistency}, Progressive Distillation~\citep{salimans2022progressive}, and adversarial approaches~\citep{sauer2024adversarial, lin2024sdxl} have provided important foundations for this line of work. Meanwhile, strong publicly released models such as Z-Image-Turbo~\citep{zimage}, Qwen-Image-Lightning~\citep{qwenimagelightning}, and FLUX.2[klein]~\citep{flux-2-2025} have demonstrated successful compression to 4--8 steps with minimal quality degradation. These results have established few-step generation as a mature and increasingly standardized stage in the model production pipeline.

A natural question arises: can we push further to 2 steps? While single-step generation remains too challenging to produce satisfactory quality, and 4--8 step approaches still leave room for efficiency gains, \textbf{2-step generation} occupies a unique position---it retains sufficient iterative structure to be leveraged, while maximizing inference efficiency. However, naively reducing the step count of existing methods to 2 leads to severe performance degradation, suggesting that the 2-step regime presents fundamentally different challenges.

Through extensive experimentation, we identify two core challenges that distinguish 2-step generation from conventional few-step distillation. The first is \textit{optimization difficulty}. With only two function evaluations, each denoising step must cover a very large interval of the noise-to-data trajectory: the first step must transform pure noise into a meaningful intermediate state, while the second step must refine this state into a clean image. In this regime, directly imposing an overly distant target distribution can make training unstable and produce persistent artifacts. We therefore find that the choice of learning target is crucial: the target should be strong enough to improve perceptual quality, but also close enough to the student's attainable distribution to provide useful gradients.

The second challenge is \textit{capacity under extreme step specialization}. In standard multi-step sampling, the same model is reused across many timesteps, and each evaluation performs a relatively local update. In contrast, the two steps of a 2-step generator play sharply different roles. This makes parameter sharing unusually restrictive: a single model must simultaneously solve two highly distinct and demanding subproblems. Moreover, because the first step determines the intermediate representation consumed by the second, optimizing the two steps independently can lead to suboptimal coordination. These observations suggest that successful 2-step distillation requires training objectives, parameterizations, and optimization procedures that are explicitly adapted to the trainability, capacity, and coordination constraints of this regime.

To address these challenges, we propose \textbf{Z-Image Turbo++}, a carefully distilled model derived from the original 8-step Z-Image Turbo and specialized for high-quality 2-step image generation. The core technical designs are as follows:
\begin{list}{$\bullet$}{\leftmargin=1.2em \itemsep=0.25em \topsep=0.1em \parsep=0pt}
    \item To make the unusually difficult 2-step objective trainable, we employ adversarial training to align our 2-step model with \textit{an existing few-step (8-step) teacher}. Critically, we find that using the few-step model's generations as real samples for the GAN discriminator---rather than external real images---provides a more stable and attainable optimization path, fundamentally improving both training stability and final quality. We attribute this to the closer distributional alignment between the teacher's and student's output, which provides more informative gradient signals.

    \item To address the insufficient effective capacity of a shared model under extreme step specialization, we propose Step-Decoupled Parameterization, where the step-specific models are initialized from the same teacher weights but updated independently thereafter, effectively enlarging model capacity and reducing interference between the two sharply different denoising tasks.

    \item We introduce end-to-end training that treats the 2-step generation process as a fully differentiable pipeline, enabling the first step to receive gradients that directly optimize the final output quality. To preserve the pretrained model's iterative generation pattern, we further retain an explicit step-1 loss as iterative regularization. This improves capacity utilization while allowing the two steps to coordinate more flexibly.

\end{list}

With these designs, our model substantially narrows the gap between 2-step and 8-step generation, preserving most of the teacher's visual quality and benchmark performance while reducing inference to only two denoising steps.

\section{Related Work}
\label{sec:related_work}

\textbf{Diffusion Model Acceleration.}
The computational cost of diffusion models has motivated extensive research on acceleration. One direction improves the sampling process through advanced ODE solvers such as DDIM~\citep{ddim}, DPM-Solver~\citep{dpmsolver}, and UniPC~\citep{unipc}, which reduce the required steps without retraining. Orthogonal approaches target the model itself through pruning~\citep{fang2023structural}, quantization~\citep{ptqd,shang2023post}, and caching~\citep{teacache, deepcache} mechanisms. Step-distillation methods, which train a student generator to reproduce a teacher's sampling behavior with fewer evaluations, are the closest to our setting and form the focus of this work.

\textbf{Few-Step Distillation.}
Progressive Distillation~\citep{salimans2022progressive} pioneered a curriculum-based approach that halves the step count iteratively. Consistency Models~\citep{song2023consistency} and their variants~\citep{wang2024phased, lu2024simplifying, ren2024hyper} enforce self-consistency along the ODE trajectory to enable direct mapping to the trajectory endpoint. Distribution Matching Distillation (DMD)~\citep{dmd, dmd2} minimizes the distributional divergence between student and teacher outputs. Decoupled DMD~\citep{decoupleddmd} clarifies the working mechanism of DMD as a CFG Augmentation engine and a Distribution Matching regularizer, enabling principled schedule design and achieving strong 4--8 step results. Other notable approaches include InstaFlow~\citep{liu2023instaflow}, Rectified Flow~\citep{liu2022flow}, and Moment Matching Distillation~\citep{mmd}.

\textbf{GAN-Based Distillation.}
Adversarial Diffusion Distillation (ADD)~\citep{sauer2024adversarial} introduced adversarial training to the distillation pipeline, using a pretrained visual feature extractor with lightweight discriminator heads. LADD~\citep{sauer2024fast} replaced this external discriminative backbone with generative features from the pretrained diffusion teacher, enabling adversarial distillation directly in latent space. SDXL-Lightning~\citep{lin2024sdxl} combines progressive distillation with adversarial objectives. DMD2~\citep{dmd2} augments the DMD loss with a GAN loss, forming a classic distillation combination that we also adopt in this work. These methods demonstrate that GAN objectives can provide powerful training signals for distillation, though the choice of real samples and training stability remain open challenges in this domain.

\textbf{Z-Image}
\citep{zimage} is a representative modern foundation model built on a Scalable Single-Stream Diffusion Transformer (S3-DiT), which processes text, image, and latent tokens within a unified transformer stream. At a 6B-parameter scale, Z-Image demonstrates strong photorealistic generation capabilities. Its distilled variant, Z-Image-Turbo, further combines Decoupled DMD~\citep{decoupleddmd} with DMDR~\citep{dmdr} to obtain an 8-step Turbo model with high visual fidelity and practical inference efficiency. Motivated by these properties, we build upon Z-Image-Turbo and study how modern single-stream diffusion transformers can be pushed into the more aggressive 2-step generation regime.

\section{Preliminary}
\label{sec:preliminary}

\subsection{Flow Matching}
\label{sec:flow_matching}

We adopt the flow matching framework~\citep{lipman2022flow} throughout this paper. We define $t=0$ as pure noise and $t=1$ as clean data. Given a data sample $x_1 \sim p_{\text{data}}$ and noise $\epsilon \sim \mathcal{N}(0, I)$, the forward process constructs an intermediate noisy sample:
\begin{equation}
    x_t = t \cdot x_1 + (1 - t) \cdot \epsilon, \quad t \in [0, 1].
\end{equation}
A neural network $v_\phi(x_t, t)$ is trained to predict the velocity field that transports the noise distribution to the data distribution. During inference, generation proceeds by integrating the learned velocity field from $t=0$ to $t=1$ using a numerical ODE solver, typically requiring many discretization steps for high quality.

\subsection{Distribution Matching Distillation}
\label{sec:dmd_review}

Distribution Matching Distillation (DMD)~\citep{dmd} trains a few-step student generator $G_\theta$ by minimizing the integral KL divergence between the student and teacher distributions:
\begin{equation}
    \mathcal{L}_{\text{IKL}}(p_{\text{real}}, p_{\text{fake}}) = \int_0^1 \mathbb{KL}(p_{\text{real}, \tau} \| p_{\text{fake}, \tau}) \, d\tau.
\end{equation}
In practice, its gradient is estimated with a frozen ``real'' score model and a concurrently trained ``fake'' score model:
\begin{equation}
\label{eq:dmd_gradient}
    \nabla_\theta \mathcal{L}_{\text{DMD}} = \mathbb{E}_{z_t, \tau, x_\tau} \left[ -\left( s^{\text{real}}_{\text{cfg}}(x_\tau) - s^{\text{fake}}_{\text{cond}}(x_\tau) \right) \frac{\partial G_\theta(z_t)}{\partial \theta} \right],
\end{equation}
where $x_\tau$ is obtained by renoising $G_\theta(z_t)$ to noise level $\tau$. This practical objective differs from the theoretical estimator, which should have used $s^{\text{real}}_{\text{cond}}$ instead of the CFG-guided score $s^{\text{real}}_{\text{cfg}}$. Despite this mismatch, CFG is crucial for strong performance in large-scale text-to-image distillation.

Decoupled DMD~\citep{decoupleddmd} explains this mismatch by decomposing the CFG-guided score difference into two terms:
\begin{equation}
\label{eq:dmd_decouple}
    s^{\text{real}}_{\text{cfg}}(x_\tau) - s^{\text{fake}}_{\text{cond}}(x_\tau)
    = 
    \alpha \left(s^{\text{real}}_{\text{cond}}(x_\tau) - s^{\text{real}}_{\text{uncond}}(x_\tau)\right)
    +
    \left(s^{\text{real}}_{\text{cond}}(x_\tau) - s^{\text{fake}}_{\text{cond}}(x_\tau)\right)
    =
    \alpha\Delta_{\text{CA}} + \Delta_{\text{DM}}.
\end{equation}
Here $\alpha$ is the CFG scale. Decoupled DMD shows that \textit{CFG Augmentation} (CA) is the main engine for few-step conversion, while \textit{Distribution Matching} (DM) mainly regularizes training and suppresses artifacts. This insight inspires principled schedule design that improves 4--8 step distillation and underlies the strong 8-step generation capability of Z-Image-Turbo.

\subsection{GAN in Distillation}
\label{sec:gan_review}

GAN objectives are widely used in diffusion distillation~\citep{sauer2024adversarial, dmd2, sauer2024fast}. A discriminator $D$ distinguishes ``real'' from ``fake'' images and provides an adversarial gradient to the generator:
\begin{equation}
    \mathcal{L}_{\text{GAN}} = \mathbb{E}_{x \sim p_{\text{real}}} [\log D(x)] + \mathbb{E}_{x \sim p_{\text{fake}}} [\log(1 - D(x))].
\end{equation}
A common practice is to freeze a cloned multi-step diffusion model as the discriminator backbone and train lightweight discriminator heads on top. We follow this architecture.

\section{Method}
\label{sec:method}

\subsection{Overview}
\label{sec:overview}

Our approach follows a two-phase pipeline:

\textbf{Phase 1: Few-Step Teacher Preparation.} We assume access to an 8-step teacher model obtained through established distillation techniques (e.g., Decoupled DMD). The production of such few-step models is well-studied and increasingly standardized in practice; we do not discuss this phase further and treat it as a given prerequisite.

\textbf{Phase 2: Two-Step Distillation.} Starting from the 8-step teacher, we distill a 2-step generator through three synergistic techniques: Distribution-Aligned Adversarial Learning (\S\ref{sec:dist_aligned}), Step-Decoupled Parameterization (\S\ref{sec:step_decoupled}), and End-to-End Training with Iterative Regularization (\S\ref{sec:e2e_training}). The overall training objective is:
\begin{equation}
    \mathcal{L} = \mathcal{L}_{\text{GAN}} + \lambda \mathcal{L}_{\text{DMD}},
\end{equation}
where $\mathcal{L}_{\text{GAN}}$ is the distribution-aligned adversarial loss and $\mathcal{L}_{\text{DMD}}$ provides complementary augmentation and regularization. Following the insights from Decoupled DMD, we set the renoising schedule as $\tau_{\text{CA}} = \tau_{\text{DM}} > t$, i.e., the renoising timestep is constrained to be \textit{cleaner} than the input timestep. We do not employ the Decoupled-Hybrid schedule ($\tau_{\text{CA}} > t, \tau_{\text{DM}} \in [0, 1]$), because in our setting the adversarial objective already provides the anti-artifact effect typically supplied by the full-range DM term. Using a shared constrained schedule also avoids the additional score-model evaluation required by Decoupled-Hybrid. We refer readers to Sec.~4.3 of~\citet{decoupleddmd} for details.

\begin{figure}[t]
    \centering
    \includegraphics[width=1\linewidth]{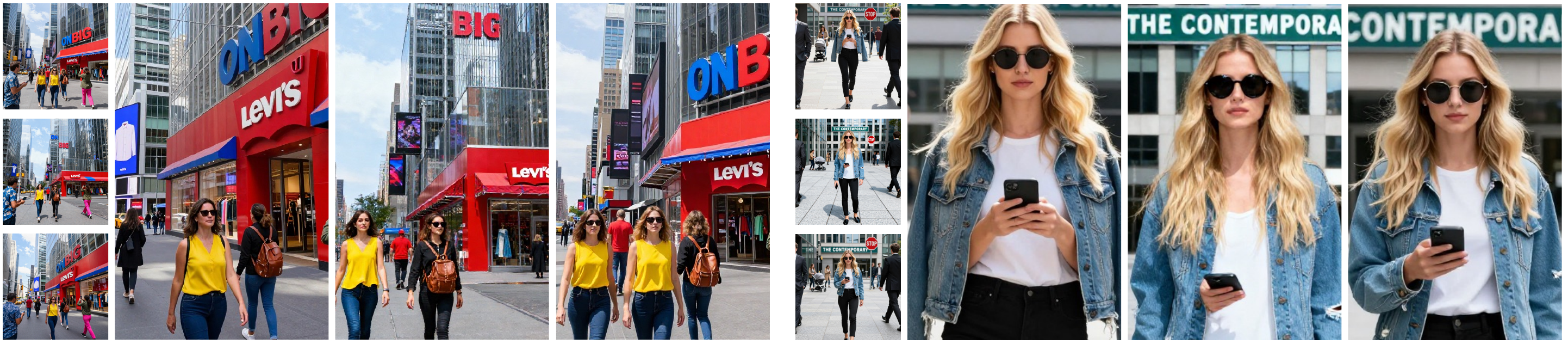}
    \caption{Adversarial training with different real-sample sources. From top to bottom and left to right: 8-step samples generated by Z-Image-Turbo, 2-step results trained with external real images, and 2-step results trained with 8-step teacher-generated images (our adopted setting).}
    \label{fig:gan_real_source}
\end{figure}

\subsection{Distribution-Aligned Adversarial Learning}
\label{sec:dist_aligned}

A key design choice in GAN-based distillation is the source of real samples for the discriminator. The conventional approach uses images from an external high-quality dataset. We challenge this default and propose using the \textbf{8-step teacher model's generated outputs} as real samples instead.

We observe that using teacher-generated images as real samples dramatically improves both training stability and final generation quality compared to using external real images. This finding is non-trivial: since real images are ``more real'' by construction, one might expect them to provide a stronger target distribution. In practice, however, using external real images leads to systematic and pronounced artifacts in the generated outputs, whereas using teacher-generated images yields stable training dynamics and clean, high-quality results.

We attribute this to distributional alignment between the real samples and the student's learning target. The 8-step teacher's output distribution is much closer to the 2-step student's target distribution than external real images. When external images are used, the discriminator can rely on distributional differences that are inherent to diffusion outputs versus natural photographs, such as texture statistics and frequency characteristics, rather than on differences that reflect generation quality. These deeply rooted differences are difficult to eliminate during post-training, and forcing the student to close them can disrupt knowledge already encoded in the model. Moreover, once such persistent cues are sufficient for the discriminator to separate real and fake samples, the discriminator has less incentive to identify more useful failure modes, reducing the effectiveness of the adversarial gradient.

Fig.~\ref{fig:loss_curves} (a) illustrates this phenomenon through training dynamics. When using 8-step teacher outputs as real samples, the generator's GAN loss exhibits a healthy pattern: it initially increases as the discriminator strengthens, then plateaus as the generator successfully closes the distribution gap. In contrast, with external real images, the loss is not only substantially higher but continues to grow throughout training, indicating an unbridgeable distributional divide that the student cannot overcome. The resulting visual difference, shown in Fig.~\ref{fig:gan_real_source}, provides direct empirical evidence for this analysis: models trained with external real images exhibit pronounced artifacts, whereas those trained with teacher-generated real samples produce clean, high-quality outputs.

This approach also fits naturally into our two-stage training paradigm. Since the 8-step teacher already has reasonable sampling speed, offline generation of teacher samples incurs manageable cost, especially compared with sampling from the original multi-step diffusion model.

\subsection{Step-Decoupled Parameterization}
\label{sec:step_decoupled}

Each denoising step in a diffusion model can be viewed as a different generation task solver. In a standard many-step sampler, the model must handle a large number of such tasks across the trajectory, but each step only covers a small interval and therefore has a relatively light local burden. Few-step models reduce the number of sampled timesteps, but each remaining step must cover a much larger interval, placing a stronger demand on the model's per-step prediction ability. This task specialization reaches its extreme in 2-step generation: the first step must construct a meaningful intermediate from near-pure noise, while the second step must turn that intermediate into a clean image.

This unusually demanding per-step requirement raises a natural question: does model capacity become a bottleneck in the 2-step regime? The favorable side of this setting is that, unlike a many-step sampler, the student only needs to handle two distinct tasks, which makes per-step parameter isolation possible. We therefore decouple the parameters for the two steps. Specifically, both steps' model weights are initialized from those of the 8-step teacher, then trained independently. This effectively doubles the model capacity dedicated to the 2-step generation task.

\begin{figure}[t]
    \centering
    \includegraphics[width=1\linewidth]{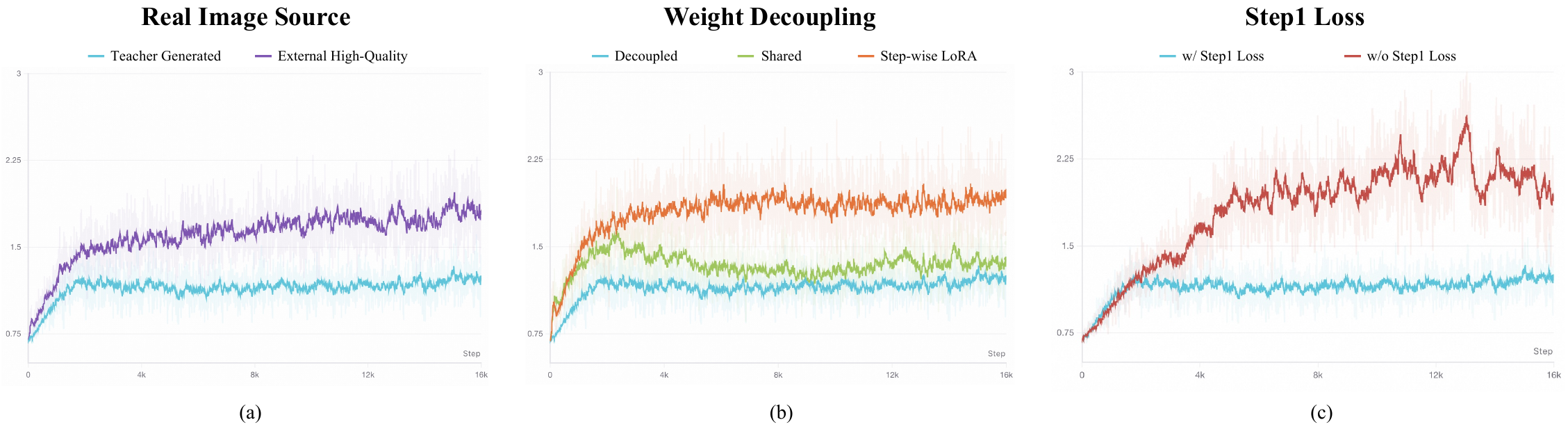}
    \caption{Generator GAN loss curves under different training settings.}
    \label{fig:loss_curves}
\end{figure}

This simple design gives rise to non-trivial improvements. As shown in Fig.~\ref{fig:loss_curves}(b), the step-2 generator GAN loss decreases substantially after parameter decoupling. Although GAN loss is affected by many factors and is not a reliable metric across substantially different experimental settings, its stable difference is informative in this controlled comparison, where the target distribution, training recipe, and optimization objective are kept fixed and only the parameterization is changed. Importantly, we also tested a weaker form of decoupling: the two steps share the same backbone but use task-specific LoRA~\citep{hu2022lora} modules. This variant performs poorly in terms of GAN loss, with a final loss even higher than full fine-tuning with a shared model. Interestingly, its benchmark performance is mixed (Tab.~\ref{tab:summary}; see Sec.~\ref{sec:experiments} for details): it improves some generic evaluation metrics, but remains clearly inferior to full weight decoupling on the most capacity-demanding and text-oriented metrics. This pattern suggests that the bottleneck arises from both fundamental model capacity and higher-level multi-task interference. Per-step LoRA can partially reduce interference through low-rank step-specific residuals, but the shared backbone must still represent two substantially different denoising maps. Full weight decoupling, by contrast, alleviates both limitations more directly.

While parameter decoupling doubles the parameter count, large-scale serving can pipeline the two step-specific models across devices, so the overall throughput and serving cost can remain nearly unchanged with proper scheduling. The trade-off is more pronounced for device-side deployment, where the additional storage can be limiting and low-memory devices may require offloading. We currently view this cost as necessary for achieving stronger 2-step quality, and leave a better balance between quality, storage, and deployment efficiency as an important future challenge.

\subsection{End-to-End Training with Iterative Regularization}
\label{sec:e2e_training}

\textbf{End-to-end training becomes feasible at 2 steps.} In multi-step diffusion models (>4 steps), end-to-end training through the entire generation chain has been considered desirable but impractical: the long computation graph leads to prohibitive memory consumption and gradient instability. The 2-step setting fundamentally changes this calculus. The entire gradient path---from initial noise through step~1 model, intermediate representation, step~2 model, to final output and loss---is concise and tractable, making full gradient tracking through both steps feasible for the first time.

End-to-end training provides two key advantages. The first is \textit{direct optimization of the first step for final quality.} Diffusion generation is a progressive process where each step determines compositional elements that subsequent steps preserve. Without end-to-end gradients, the first step can only be optimized for its local objective, potentially making choices that are locally reasonable but suboptimal for the final output. End-to-end training allows the first step to receive gradients that reflect the final generation quality, enabling it to resolve issues whose root causes originate in step~1 but only manifest after step~2. The second is \textit{flexible resource coordination}: treating both steps as a unified system allows them to implicitly coordinate their division of labor, improving overall capacity utilization.

\subsubsection{The Necessity of Step-1 Loss}
\label{sec:step1_loss}

A natural question arises: if we optimize end-to-end for the final output, is a separate loss on the first step's intermediate output still necessary? Our initial hypothesis is that removing this constraint would free the model from redundant requirements and allow full capacity allocation to final quality.

\begin{figure}[t]
    \centering
    \includegraphics[width=1\linewidth]{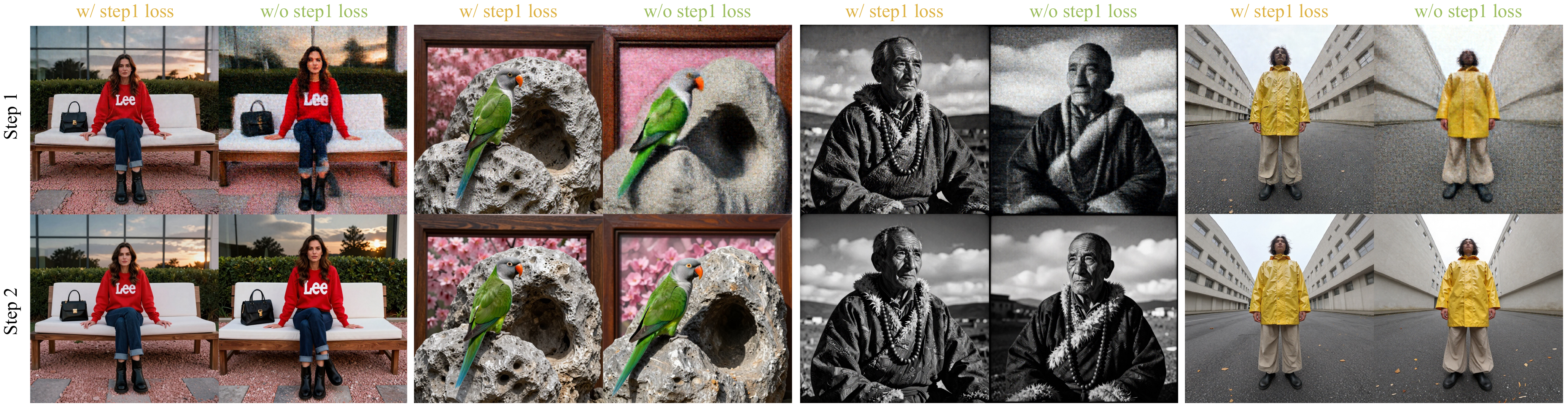}
    \caption{Visual comparison showing degraded quality when step-1 loss is removed.}
    \label{fig:step1_loss}
\end{figure}

\textit{Contrary to this intuition, experiments reveal that removing the step-1 loss causes the step-2 generator's GAN loss to surge, accompanied by visible degradation in generation quality (Fig.~\ref{fig:step1_loss}) and benchmark results (Tab.~\ref{tab:summary})}. The step-1 loss is therefore essential for stable and high-quality training.

We explain this behavior from a transfer-learning perspective. The distilled model's final performance depends on two factors: (1) learning capacity for the downstream task, and (2) the ability to leverage knowledge accumulated during pretraining. Diffusion models possess a deeply ingrained \textit{iterative nature}: they are pretrained to perform progressive denoising, where each step produces a meaningful intermediate that the next step can build upon.

Removing the step-1 loss increases learning flexibility (factor 1), but it also disrupts the iterative generation pattern that the pretrained model has internalized (factor 2). As shown in Fig.~\ref{fig:step1_loss}, the intermediate output degenerates into a low-quality representation rather than a meaningful partial generation. Such a representation might be viable for a model trained from scratch, but it is poorly aligned with the inductive biases encoded in the pretrained weights. By maintaining step-1 generation quality through an explicit loss, we help the model \textit{preserve its familiar iterative generation mode}, enabling more effective transfer of pretrained knowledge. Our experiments indicate that, in this regime, transfer efficiency is more critical than unconstrained learning flexibility.

\subsubsection{Important Implementation Details}
\label{sec:impl_details}

Our implementation keeps the desired end-to-end training signal while avoiding the peak memory cost of a fully connected two-step graph. The key observation is that [first-step local loss] and [second-step $\rightarrow$ final loss] form two independent branches after the first-step velocity prediction. We therefore first propagate through the second branch through a detached clone of first-step prediction, store the resulting gradient, and then inject it into the first-step model by an \textit{inherit loss} that adds to step~1's original local loss. This avoids the memory occupation of the first branch during the propagation of the second branch, thus reducing peak memory requirement. In practice, we scale the inherited term by a small weight (0.1) to prevent gradient explosion. Together with per-transformer-block gradient checkpointing and FSDP, this keeps the overall memory cost within a practical range. Appendix~\ref{sec:app_impl_pseudocode} provides the corresponding pseudo-code.

\begin{figure}[t]
    \centering
    \includegraphics[width=1\linewidth]{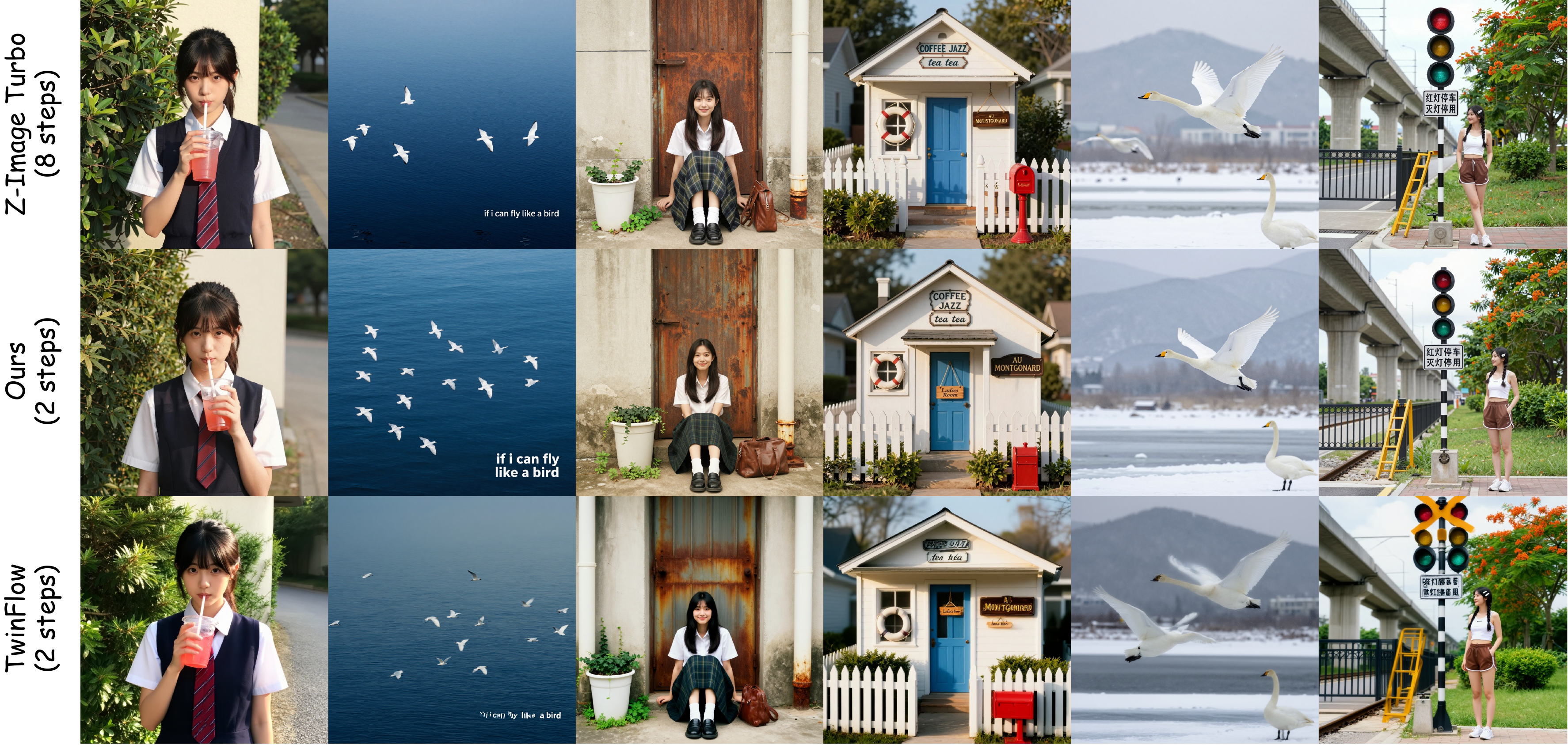}
    \caption{Qualitative comparison among 8-step Z-Image-Turbo, TwinFlow, and our Z-Image Turbo++. Our method achieves a better quality-efficiency trade-off.}
    \label{fig:qualitative}
\end{figure}

\begin{table}[t]
\centering
\caption{Overall performance comparison across benchmarks.}
\label{tab:summary}
\resizebox{\linewidth}{!}{%
\begin{tabular}{llcccccc}
\toprule
Idx & Model & NFE & OneIG & GenEval & DPG-Bench & LongText-CN & LongText-EN \\
\midrule
\firsttablegroup{8}{Component Ablation}
\circnum{1} & Baseline & 2 & 51.72 & \textbf{77.50} & 84.74 & 87.13 & 83.29 \\
\circnum{2} & \ding{192} + Teacher as Real & 2 & 51.89 & 72.72 & 84.86 & 87.16 & {\ul 86.03} \\
\circnum{3} & \ding{193} + Decoupling Weight & 2 & {\ul 52.15} & 74.67 & {\ul 85.39} & {\ul 89.27} & 85.81 \\
\circnum{4} & \ding{194} + End-to-End Training (Ours) & 2 & \textbf{52.50} & {\ul 75.70} & \textbf{85.86} & \textbf{91.62} & \textbf{89.88} \\
\addlinespace[1pt]
\tablegroup{8}{Weight Decoupling Ablation}
\circnum{4} & Ours & 2 & \textbf{52.50} & {\ul 75.70} & {\ul 85.86} & \textbf{91.62} & \textbf{89.88} \\
\circnum{5} & \ding{195} w/ Shared Weight & 2 & 50.67 & 73.62 & 85.51 & {\ul 87.64} & {\ul 81.14} \\
\circnum{6} & \ding{195} w/ Per-Step LoRA & 2 & {\ul 50.96} & \textbf{76.10} & \textbf{86.58} & 80.71 & 76.90 \\
\addlinespace[1pt]
\tablegroup{8}{Step-1 Loss Ablation}
\circnum{4} & Ours & 2 & \textbf{52.50} & \textbf{75.70} & \textbf{85.86} & \textbf{91.62} & \textbf{89.88} \\
\circnum{7} & \ding{195} w/o Step-1 Loss & 2 & {\ul 50.16} & {\ul 71.02} & {\ul 83.98} & {\ul 84.49} & {\ul 82.07} \\
\addlinespace[1pt]
\tablegroup{8}{Comparative}
\circnum{4} & Ours & 2 & \textbf{52.50} & 75.70 & {\ul 85.86} & \textbf{91.62} & \textbf{89.88} \\
\circnum{8} & Twin Flow & 2 & {\ul 51.38} & 72.41 & \textbf{85.98} & 78.65 & 71.99 \\
\circnum{9} & DMD2 & 2 & 50.70 & {\ul 76.12} & 85.40 & {\ul 85.30} & {\ul 80.99} \\
\circnum{10} & Z-Image-Turbo & 2 & 50.94 & \textbf{76.53} & 85.78 & 78.48 & 72.89 \\
\addlinespace[1pt]
\tablegroup{8}{Reference}
\circnum{11} & Z-Image-Turbo & 8 & 52.84 & 75.01 & 84.86 & 92.56 & 91.74 \\
\bottomrule
\end{tabular}
}
\end{table}

\begin{table}[t]
\centering
\caption{Detailed results on OneIG benchmark.}
\label{tab:oneig_detail}
\resizebox{\linewidth}{!}{%
\begin{tabular}{llccccccc}
\toprule
Idx & Model & NFE & Overall & Alignment & Text & Diversity & Style & Reasoning \\
\midrule
\firsttablegroup{9}{Component Ablation}
\circnum{1} & Baseline & 2 & 51.72 & {\ul 85.10} & 94.79 & \textbf{13.14} & 36.93 & 28.65 \\
\circnum{2} & \ding{192} + Teacher as Real & 2 & 51.89 & 85.00 & 96.46 & 12.70 & 36.14 & {\ul 29.16} \\
\circnum{3} & \ding{193} + Decoupling Weight & 2 & {\ul 52.15} & 84.98 & {\ul 96.49} & {\ul 13.05} & {\ul 37.18} & 29.07 \\
\circnum{4} & \ding{194} + End-to-End Training (Ours) & 2 & \textbf{52.50} & \textbf{85.65} & \textbf{97.09} & 11.75 & \textbf{37.87} & \textbf{30.12} \\
\addlinespace[1pt]
\tablegroup{9}{Weight Decoupling Ablation}
\circnum{4} & Ours & 2 & \textbf{52.50} & \textbf{85.65} & \textbf{97.09} & {\ul 11.75} & \textbf{37.87} & \textbf{30.12} \\
\circnum{5} & \ding{195} w/ Shared Weight & 2 & 50.67 & 84.68 & {\ul 92.35} & 11.10 & 36.79 & {\ul 28.44} \\
\circnum{6} & \ding{195} w/ Per-Step LoRA & 2 & {\ul 50.96} & {\ul 85.61} & 91.20 & \textbf{12.54} & {\ul 37.32} & 28.12 \\
\addlinespace[1pt]
\tablegroup{9}{Step-1 Loss Ablation}
\circnum{4} & Ours & 2 & \textbf{52.50} & \textbf{85.65} & \textbf{97.09} & \textbf{11.75} & \textbf{37.87} & \textbf{30.12} \\
\circnum{7} & \ding{195} w/o Step-1 Loss & 2 & {\ul 50.16} & {\ul 82.70} & {\ul 94.69} & {\ul 11.46} & {\ul 33.67} & {\ul 28.27} \\
\addlinespace[1pt]
\tablegroup{9}{Comparative}
\circnum{4} & Ours & 2 & \textbf{52.50} & \textbf{85.65} & \textbf{97.09} & 11.75 & {\ul 37.87} & \textbf{30.12} \\
\circnum{8} & Twin Flow & 2 & {\ul 51.38} & {\ul 85.26} & 90.61 & \textbf{15.49} & \textbf{38.03} & 27.52 \\
\circnum{9} & DMD2 & 2 & 50.70 & 85.21 & 90.82 & 12.75 & 36.77 & {\ul 27.93} \\
\circnum{10} & Z-Image-Turbo & 2 & 50.94 & 83.92 & {\ul 92.85} & {\ul 13.85} & 36.25 & 27.86 \\
\addlinespace[1pt]
\tablegroup{9}{Reference}
\circnum{11} & Z-Image-Turbo & 8 & 52.84 & 84.03 & 99.32 & 13.85 & 36.81 & 30.19 \\
\bottomrule
\end{tabular}
}
\end{table}

\section{Experiments}

\label{sec:experiments}

\textbf{Qualitative Results.} Fig.~\ref{fig:overview} presents representative images generated by Z-Image Turbo++ using only two inference steps. Despite this extremely compressed sampling budget, the model produces rich details, sharp textures, and strong text-rendering ability. In particular, it preserves most of the image quality and photorealistic appearance for which the original Z-Image model is known. Fig.~\ref{fig:qualitative} further compares our method against two important baselines: the original 8-step Z-Image Turbo model and TwinFlow, a recent 2-step distillation method. Visually, both Z-Image Turbo++ and TwinFlow retain basic generation capability, but our model shows clear advantages in three aspects: better preservation of global coherence and realistic style, more faithful reproduction of high-frequency details, and fewer systematic artifacts. The difference is especially pronounced in text generation, a challenging and discriminative dimension for ultra-few-step compression: TwinFlow exhibits substantial degradation, whereas our model maintains considerably higher text quality.

\textbf{Quantitative Results.} We evaluate on four standard benchmarks: DPGBench~\citep{dpgbench}, GenEval~\citep{geneval}, OneIGBench~\citep{oneig}, and LongTextBench~\citep{longtext}. All prompts are taken directly from the official benchmark sets without prompt enhancement. The main results are summarized in Table~\ref{tab:summary}. Since OneIGBench provides the most comprehensive coverage, we report its detailed breakdown in Table~\ref{tab:oneig_detail}; detailed results for the remaining benchmarks are provided in the appendix.

Across the full table, our method (\circnum{4}) achieves strong results on all four benchmarks, but the trends are not uniform across benchmarks. OneIGBench and LongTextBench show clearer separation among model variants, whereas GenEval and DPGBench are more mixed; for example, the per-step LoRA variant (\circnum{6}) outperforms the 8-step teacher (\circnum{11}) on these two benchmarks despite being much worse on OneIGBench and LongTextBench. We therefore interpret the results jointly across benchmarks, leaving a detailed study of metric divergence to future work.

Overall, our model outperforms TwinFlow (using its officially released checkpoint), our reimplementation of DMD2, and direct 2-step inference with Z-Image Turbo, while approaching the 8-step Turbo baseline on nearly all metrics. Nevertheless, a small but consistent gap remains in text generation, as reflected by LongText-CN, LongText-EN, and OneIG-Text. This is consistent with our qualitative observations: the model largely matches the 8-step baseline in visual fidelity and primary object generation, but remains less reliable for dense text and secondary or underspecified objects in complex scenes. These limitations point to an important direction for future work.

\textbf{Ablation summary.} For the GAN target distribution, using external high-quality data as real samples leads to training instability and systematic artifacts, as shown in Fig.~\ref{fig:gan_real_source}. Replacing them with 8-step teacher-generated images removes these artifacts and yields visually more natural outputs. For parameter decoupling, assigning independent weights to the two steps not only lowers the generator GAN loss, suggesting that the 2-step distribution becomes harder to distinguish from the 8-step teacher distribution, but also improves quantitative performance. Importantly, the comparison between \circnum{3} and \circnum{5} in Tab.~\ref{tab:summary} suggests that end-to-end training relies more heavily on weight decoupling, likely because the two steps require stronger functional specialization when they are optimized jointly. Meanwhile, the per-step LoRA variant still falls short of full weight decoupling, especially on difficult text-generation benchmarks, supporting our hypothesis that 2-step generation is constrained by a capacity bottleneck. Finally, we show that even under end-to-end training, a separate step-1 loss remains indispensable (Fig.~\ref{fig:step1_loss}). Without it, the step-1 intermediate output degenerates into a low-quality representation, breaking the iterative refinement pattern deeply embedded in diffusion models and leading to unstable training and degraded final performance.

\section{Conclusion and Limitations}
\label{sec:conclusion}

In this work, we introduce \textbf{Z-Image Turbo++}, a 2-step image generation model distilled from the 8-step Z-Image Turbo teacher. By combining Distribution-Aligned Adversarial Learning, Step-Decoupled Parameterization, and End-to-End Training with Iterative Regularization, our method addresses the stability, capacity, and knowledge-preservation challenges of the 2-step regime. The main limitation is additional parameter storage. Besides, while our model substantially narrows the gap between 2-step and 8-step generation, challenging cases such as dense text rendering, secondary objects, and complex-scene generation remain less reliable than the 8-step teacher. Finally, our end-to-end training framework may provide a useful foundation for reinforcement learning-based optimization of few-step generators, but we leave this direction to future work.

\newpage

{
\small
\bibliography{ultraturbo}
\bibliographystyle{plainnat}
}

\newpage
\appendix
\section{Pseudo-code for Memory-Efficient Training}
\label{sec:app_impl_pseudocode}

\begin{lstlisting}[style=pythoncode, caption={Pseudo-code for the naive two-step generator update and our memory-efficient implementation with inherited gradients.}, label={lst:memory_efficient_training}]
def naive_generator_training(step0_model, step1_model,
                             step0_stride, step1_stride,
                             caption, noise):
    assert step0_stride + step1_stride == 1.0

    step0_v_prediction = step0_model(
        noise, caption, time_cond=0.0)
    step0_x_prediction = noise + step0_v_prediction * 1.0

    step1_input = noise + step0_v_prediction * step0_stride
    step1_v_prediction = step1_model(
        step1_input, caption, time_cond=step0_stride)
    step1_x_prediction = step1_input + \
        step1_v_prediction * step1_stride

    # Loss computation also requires teacher-generated images
    # and other training states, omitted with ... for simplicity.
    step0_loss = GAN_LOSS(step0_x_prediction, ...) + \
                 DMD_LOSS(step0_x_prediction, ...)
    step1_loss = GAN_LOSS(step1_x_prediction, ...) + \
                 DMD_LOSS(step1_x_prediction, ...)

    (step0_loss + step1_loss).backward()

def memory_efficient_generator_training(
        step0_model, step1_model, step0_stride, step1_stride,
        caption, noise, inherit_weight=0.1):
    assert step0_stride + step1_stride == 1.0

    step0_v_prediction = step0_model(
        noise, caption, time_cond=0.0)

    # First handle the path from step0_v_prediction to step1_loss.
    # detach() cuts the graph here; the gradient is later
    # passed back to step0_model through inherit_loss.
    step0_v_detached = step0_v_prediction.clone().detach()
    step0_v_detached.requires_grad_(True)

    step1_input = noise + step0_v_detached * step0_stride
    step1_v_prediction = step1_model(
        step1_input, caption, time_cond=step0_stride)
    step1_x_prediction = step1_input + \
        step1_v_prediction * step1_stride

    step1_loss = GAN_LOSS(step1_x_prediction, ...) + \
                 DMD_LOSS(step1_x_prediction, ...)
    step1_loss.backward()

    # Then handle step0's own loss and inherited step1 gradient.
    step0_x_prediction = noise + step0_v_prediction * 1.0
    inherit_loss = torch.sum(
        step0_v_prediction * step0_v_detached.grad.detach())

    step0_loss = GAN_LOSS(step0_x_prediction, ...) + \
                 DMD_LOSS(step0_x_prediction, ...)

    # Directly passing the full gradient to step0_model can
    # lead to gradient explosion; inherit_weight modulates it.
    total_step0_loss = step0_loss + inherit_weight * inherit_loss
    total_step0_loss.backward()
\end{lstlisting}

\clearpage
\section{More Experimental Details}
\label{sec:result_details}

\subsection{Experiment Settings}
\label{sec:app_experiment_settings}

For all trainable models, we use the Adam optimizer with a learning rate of $1\times10^{-5}$, $\beta_1=0.0$, $\beta_2=0.9$, and no weight decay. Following the TTUR strategy in DMD2~\citep{dmd2}, we update the generator and the guidance model with a frequency ratio of $1{:}5$. The full training runs for $20{,}000$ iterations, and we maintain an exponential moving average of the generator parameters with a decay rate of $0.99$.

Before training, we pre-generate samples from the 8-step teacher and use them as the teacher distribution for subsequent training. All experiments are conducted on 16 H100 GPUs with a global batch size of 64. Under our implementation, the complete training process takes approximately 80 hours.

For the loss weights, we set both the DMD loss on the generator and the corresponding diffusion loss on the guidance model to $1\times10^{-2}$. The GAN generator loss and discriminator loss are both weighted by $1\times10^{-3}$, and the inherit loss weight is set to $0.1$.

\begin{table}[h]
\centering
\caption{Detailed results on DPG-Bench benchmark.}
\label{tab:dpgbench_detail}
\resizebox{\linewidth}{!}{%
\begin{tabular}{llccccccc}
\toprule
Idx & Model & NFE & overall & attribute & global & relation & other & entity \\
\midrule
\firsttablegroup{9}{Component Ablation}
\circnum{1} & Baseline & 2 & 84.74 & 89.08 & 83.28 & 93.62 & 86.40 & 91.10 \\
\circnum{2} & \ding{192} + Teacher as Real & 2 & 84.86 & 88.90 & 82.98 & {\ul 93.73} & 87.60 & 91.11 \\
\circnum{3} & \ding{193} + Decoupling Weight & 2 & {\ul 85.39} & \textbf{90.47} & \textbf{91.32} & 91.03 & \textbf{89.83} & {\ul 91.58} \\
\circnum{4} & \ding{194} + End-to-End Training (Ours) & 2 & \textbf{85.86} & {\ul 89.30} & {\ul 83.89} & \textbf{94.58} & {\ul 89.60} & \textbf{91.84} \\
\addlinespace[1pt]
\tablegroup{9}{Weight Decoupling Ablation}
\circnum{4} & Ours & 2 & {\ul 85.86} & \textbf{89.30} & {\ul 83.89} & {\ul 94.58} & \textbf{89.60} & {\ul 91.84} \\
\circnum{5} & \ding{195} w/ Shared Weight & 2 & 85.51 & 88.88 & 82.07 & 94.00 & {\ul 89.20} & 91.31 \\
\circnum{6} & \ding{195} w/ Per-Step LoRA & 2 & \textbf{86.58} & {\ul 89.26} & \textbf{84.80} & \textbf{94.89} & 86.80 & \textbf{92.37} \\
\addlinespace[1pt]
\tablegroup{9}{Step-1 Loss Ablation}
\circnum{4} & Ours & 2 & \textbf{85.86} & \textbf{89.30} & \textbf{83.89} & \textbf{94.58} & \textbf{89.60} & \textbf{91.84} \\
\circnum{7} & \ding{195} w/o Step-1 Loss & 2 & {\ul 83.98} & {\ul 88.28} & {\ul 80.85} & {\ul 93.08} & {\ul 86.40} & {\ul 90.34} \\
\addlinespace[1pt]
\tablegroup{9}{Comparative}
\circnum{4} & Ours & 2 & {\ul 85.86} & \textbf{89.30} & {\ul 83.89} & \textbf{94.58} & \textbf{89.60} & \textbf{91.84} \\
\circnum{8} & Twin Flow & 2 & \textbf{85.98} & {\ul 89.02} & 81.46 & {\ul 94.16} & {\ul 87.60} & {\ul 91.83} \\
\circnum{9} & DMD2 & 2 & 85.40 & 88.88 & 83.59 & 93.81 & 85.60 & 91.10 \\
\circnum{10} & Z-Image-Turbo & 2 & 85.78 & 88.78 & \textbf{84.19} & {\ul 94.16} & \textbf{89.60} & 91.73 \\
\addlinespace[1pt]
\tablegroup{9}{Reference}
\circnum{11} & Z-Image-Turbo & 8 & 84.86 & 90.14 & 91.29 & 92.16 & 88.68 & 89.59 \\
\bottomrule
\end{tabular}
}
\end{table}

\begin{table}[h]
\centering
\caption{Detailed results on GenEval benchmark.}
\label{tab:geneval_detail}
\resizebox{\linewidth}{!}{%
\begin{tabular}{llcccccccc}
\toprule
Idx & Model & NFE & overall & two\_object & color\_attr & position & counting & colors & single\_object \\
\midrule
\firsttablegroup{10}{Component Ablation}
\circnum{1} & Baseline & 2 & \textbf{77.50} & \textbf{88.64} & \textbf{67.75} & \textbf{51.25} & \textbf{70.94} & \textbf{86.44} & \textbf{100.00} \\
\circnum{2} & \ding{192} + Teacher as Real & 2 & 72.72 & {\ul 87.63} & 61.50 & 43.50 & 57.81 & {\ul 85.90} & \textbf{100.00} \\
\circnum{3} & \ding{193} + Decoupling Weight & 2 & 74.67 & 87.12 & 62.75 & 46.00 & 68.12 & 84.04 & \textbf{100.00} \\
\circnum{4} & \ding{194} + End-to-End Training (Ours) & 2 & {\ul 75.70} & {\ul 87.63} & {\ul 64.50} & {\ul 48.75} & {\ul 69.06} & 84.57 & {\ul 99.69} \\
\addlinespace[1pt]
\tablegroup{10}{Weight Decoupling Ablation}
\circnum{4} & Ours & 2 & {\ul 75.70} & {\ul 87.63} & \textbf{64.50} & {\ul 48.75} & 69.06 & {\ul 84.57} & {\ul 99.69} \\
\circnum{5} & \ding{195} w/ Shared Weight & 2 & 73.62 & 84.09 & {\ul 61.25} & 44.00 & {\ul 70.00} & 82.98 & 99.38 \\
\circnum{6} & \ding{195} w/ Per-Step LoRA & 2 & \textbf{76.10} & \textbf{90.40} & 59.25 & \textbf{50.00} & \textbf{71.56} & \textbf{85.37} & \textbf{100.00} \\
\addlinespace[1pt]
\tablegroup{10}{Step-1 Loss Ablation}
\circnum{4} & Ours & 2 & \textbf{75.70} & \textbf{87.63} & \textbf{64.50} & \textbf{48.75} & \textbf{69.06} & \textbf{84.57} & \textbf{99.69} \\
\circnum{7} & \ding{195} w/o Step-1 Loss & 2 & {\ul 71.02} & {\ul 82.83} & {\ul 61.25} & {\ul 45.50} & {\ul 54.69} & {\ul 82.18} & \textbf{99.69} \\
\addlinespace[1pt]
\tablegroup{10}{Comparative}
\circnum{4} & Ours & 2 & 75.70 & 87.63 & {\ul 64.50} & 48.75 & {\ul 69.06} & 84.57 & {\ul 99.69} \\
\circnum{8} & Twin Flow & 2 & 72.41 & 87.12 & 54.75 & 40.00 & 67.81 & {\ul 85.11} & {\ul 99.69} \\
\circnum{9} & DMD2 & 2 & {\ul 76.12} & \textbf{90.40} & \textbf{65.50} & {\ul 49.00} & 67.50 & 84.31 & \textbf{100.00} \\
\circnum{10} & Z-Image-Turbo & 2 & \textbf{76.53} & {\ul 89.39} & 57.75 & \textbf{50.50} & \textbf{74.38} & \textbf{87.77} & 99.38 \\
\addlinespace[1pt]
\tablegroup{10}{Reference}
\circnum{11} & Z-Image-Turbo & 8 & 75.01 & 88.89 & 58.75 & 45.75 & 71.88 & 85.11 & 99.69 \\
\bottomrule
\end{tabular}
}
\end{table}

\end{document}